# A Nascent Taxonomy of Machine Learning in Intelligent Robotic Process Automation

Lukas Laakmann[0009−0007−8076−1385], Seyyid A. Ciftci[0009−0000−7840−4644], and Christian Janiesch[0000−0002−8050−123X]

TU Dortmund University, Dortmund, Germany
{lukas.laakmann|seyyid.ciftci|christian.janiesch}@tu-dortmund.de

**Abstract.** Robotic process automation (RPA) is a lightweight approach to automating business processes using software robots that emulate user actions at the graphical user interface level. While RPA has gained popularity for its cost-effective and timely automation of rule-based, well-structured tasks, its symbolic nature has inherent limitations when approaching more complex tasks currently performed by human agents. Machine learning concepts enabling intelligent RPA provide an opportunity to broaden the range of automatable tasks. In this paper, we conduct a literature review to explore the connections between RPA and machine learning and organize the joint concept intelligent RPA into a taxonomy. Our taxonomy comprises the two meta-characteristics RPA-ML integration and RPA-ML interaction. Together, they comprise eight dimensions: architecture and ecosystem, capabilities, data basis, intelligence level, and technical depth of integration as well as deployment environment, lifecycle phase, and user-robot relation.

## 1 Introduction

In recent years, the field of process automation has undergone a remarkable transformation, with robotic process automation (RPA) revolutionizing how businesses across various industries manage and streamline their operations in a timely fashion. RPA, with its ability to automate repetitive, rule-based tasks, has emerged as a crucial tool for enhancing efficiency, reducing costs, and improving overall productivity [40]. Despite its widespread market acceptance, RPA has limitations. Traditional RPA robots are confined to executing repetitive rule-based processes as configured by recordings or flowcharts. They lack the flexibility to adapt to changes in context or process unstructured data [36].

Concurrently, machine learning (ML) has witnessed unprecedented advancements, offering powerful capabilities for example in data analysis, pattern recognition, and text generation [19,11]. The integration of RPA and ML presents a logical next step, promising more comprehensive and intelligent levels of automation in business processes [17], for example to enable sharing data among multiple parties within neutral data trust models.

As both, RPA and ML technologies, continue to evolve and possibly converge, a need arises for a systematic taxonomy that can effectively categorize and delineate the multifaceted landscape of RPA implementations that harness ML concepts. The absence of a standardized frame for discussion hinders not only the clarity of communication within and across the RPA and ML communities but also the effective evaluation, comparison, and selection of RPA solutions by organizations seeking to intelligentize their automation journey. By providing a structured and standardized vocabulary, our taxonomy will enable practitioners, researchers, and organizations to communicate more precisely about their RPA-ML initiatives, fostering a deeper understanding of the field's potential and limitations.

With this article, we seek to address this gap in the literature by advocating for the development of a nascent taxonomy for RPA with ML integration. It will help to systematize the new and substantive developments in RPA and ML and – foremost – it synthesizes knowledge from different lines of research. It may even assist in identifying missing or neglected themes. Consequently, we pose our research question as follows:

**RQ:** *"What are the dimensions and characteristics of a taxonomy of RPA and ML integration and use?"*

To answer this question, we conducted a hermeneutic literature review [5] and developed a taxonomy based on a sample of 45 publications following a iterative taxonomy development process [29]. The resulting nascent taxonomy is useful, comprehensive, extensible, and explaining for IS researchers as well as for practitioners.

Our paper is organized as follows: In Section 2, we introduce common concepts of RPA and ML separately. Section 3 contains an outline of our research methodology. We present the taxonomy in Section 4 before we close with a brief discussion and summary.

## 2   Background and Related Literature

### 2.1   Robotic Process Automation

We characterize RPA as an umbrella term for automation technology operating on the user interface (UI) level, evolving from the traditional screen scraper method that recorded and replicated precise user mouse interactions. It uses software robots to mimic human interaction with legacy software so that there are no adjustments necessary to existing software [17]. RPA addresses the limitation of UI shifts by processing user interactions at a logical UI element level, rather than based on coordinates [31]. RPA also follows in the tradition of older lightweight automation methods like macros or scripting, which consolidate individual user interactions .

RPA is particularly suitable for mundane tasks where humans extract structured data from one or more applications, process it rule-based, and re-enter it into another system, known as "swivel-chair processes" [24]. The goal is to relieve users of such tasks, allowing them to focus on higher-value activities [25].

In contrast to other traditional automation techniques, RPA operates "outside-in", that is it builds upon the existing information system architecture without altering it. This allows RPA to connect various applications without the for need for an explicit application programming interface (API) but the UI [40]. Despite this technical difference, RPA also changes the organizational and social aspects of automation by allowing user-side RPA robot development via low-code flowcharts outside a dedicated project in the IT department [36].

In a nutshell, symbolic RPA suits rule-based, non-interactive tasks and is not ideal for dynamic environments needing flexibility [27]. It is limited by structured data requirements and may overlook exceptional cases due to "happy path" recording. In the context of intelligent RPA, especially technical limitations of common symbolic RPA shall be addressed through ML integration.

### 2.2 Machine Learning

Various nuances of definitions for the terms artificial intelligence (AI) and ML have been identified. Common to all these approaches is the definition of AI as the ability of machines to solve cognitively human-like tasks [4]. For instance, Haenlein and Kaplan [15] define AI more specifically as the ability of a system to accurately interpret external data, learn from it, and apply the learned knowledge through flexible adaptation to specific tasks and objectives.

ML is one method that machines can potentially employ to achieve AI. While early approaches like expert systems required rules and axioms to be explicitly defined and used logic to infer knowledge, ML autonomously generates knowledge from training data through experience, eliminating the need for explicit programming [19]. Its current state of development can be characterized as *narrow artificial superintelligence* that can exhibit super-human capabilities for specific tasks that it was selectively trained for such as playing board games or analyzing medical scans.

In the field of ML, distinctions can be made based on algorithmic learning types, tasks, and model architectures. Depending on the data provided to the learning model, we distinguish three learning types [19] : supervised learning, unsupervised learning, and reinforcement learning. Furthermore, ML algorithms can be differentiated based on their structure as shallow or deep learning [19].

### 2.3 Related Work on Intelligent RPA

The literature provides numerous categorizations of how RPA and ML can be combined [1,28,31,38,41]. However, these categorizations often focus on specific aspects and do not provide a comprehensive differentiation. To gain a complete understanding of the complexity of the possibilities for enhancing RPA with ML, it is necessary to interconnect and relate these various dimensions.

On the other hand, not all dimensions found in the literature offer relevant insights into the conceptual integration with ML. For example, the distinction between whether UI recorders operate at the browser, system, or application level is merely descriptive of RPA, regardless of whether it is enriched with ML [1]. Similarly, the literature does not always differentiate between the specific ML algorithms used, as it is more important in the context of RPA to focus on their capabilities and general functionality. Technical implementation details are typically not emphasized in the literature because the use in the RPA context is a transfer application.

Our undertaking is further complicated by the widespread use of the term AI in industry, often applied to all innovative approaches that are marketed as AI regardless of the actual intelligence involved. Competing definitions must also be reconciled. For example, while Beerbaum [3] attributes the same meaning to *smart process automation*, cognitive RPA, and intelligent RPA in the sense of combining RPA with AI, Lacity and Willcocks [25] explain that intelligent RPA is the convergence of RPA and cognitive automation (CA), where CA broadly signifies automation through AI.

Further, Lacity and Willcocks [25] argue that a strict one-dimensional separation between forms like symbolic RPA and CA is not the right perspective. Instead, we observe a continuum, where RPA is gradually enriched by cognitive elements. We address this finding by elaborating different levels of ML integration and the complexity of integration approaches through a multidimensional characterization.

Given the nascent state of the research field, it is expected that terminologies will further evolve but become more standardized over time. Our nascent taxonomy can help to shape this journey.

## 3   Research Design and Methodological Approach

We conducted an integrative literature review to explore concepts for structuring the connection between RPA and ML employing a hermeneutic approach [5]. We searched multiple times for relevant literature in the databases *AIS eLibrary*, *IEEE Xplore*, *ACM Digital Library*, *SpringerLink*, and *ScienceDirect* with an evolving search string eventually resembling: *("AI" OR "ML" OR "Artificial Intelligence" OR "Machine Learning") AND ("RPA" OR "Robotic Process Automation")*.

This search string takes into account the insight gained in the preliminary research that there is often no clear differentiation between the terms AI and ML. Therefore, limiting the search to the term ML might omit relevant publications. Furthermore, the search results were narrowed down according to the capabilities of each database and filtered. Due to the lack of scientific literature, as mentioned in overview articles such as those by Syed et al. [36] or Chugh et al. [7], there was a deliberate choice not to further restrict to high-quality journals or similar sources. Publications in languages other than English or German were not found during the search and, therefore, did not need to be explicitly excluded. Subsequently, we conducted an initial review of the search results based on titles and abstracts. We excluded publications that were not research articles from computer science or information systems, did not explicitly link RPA and ML in the title or the abstract, adopted a highly individual perspective, focused on other AI concepts than ML or concentrated only on implications, but lacked a conceptual perspective on connecting RPA and ML.

This process resulted in a reduction to 51 potential articles. The remaining publications were analyzed following the methodology outlined by Boell and Cecez-Kecmanovic [5], and they were classified based on their central ideas and methodologies. After removing duplicates, a backward and forward search was conducted. As suggested by Boell and Cecez-Kecmanovic, literature searches may never truly conclude, but at some point saturation is reached. In this case, that point of saturation occurred with 45 publications that we reviewed in depth. Therefore, this literature review can only be considered representative and not exhaustive.

To gain a structure of the discovered concepts, we applied the taxonomy development method proposed by Nickerson et al. [29] that shall avoid ad hoc classification by iterating and defined ending conditions. As meta-characteristic, we choose conceptual properties of *structure* and *process* to specify how ML augments RPA: *RPA-ML integration* to represent structural topics and *RPA-ML interaction* to represent the augmentation in use.

In the present case, we strived to develop a taxonomy based on literature sources. However, the method by Nickerson et al. [29] is designed for synthesis based on individual exemplars. After a fundamental analysis of the selected publications for the review, it was observed that many of the publications ($n = 26$) not only discuss individual instances but also provide a framework or conceptual differentiations for connecting RPA with ML. Therefore, we organized the approach as follows: first we collected dimensions and characteristics based on these works in the conceptual-deductive sub-circle, and then we analyzed the instances in application-oriented publications using the empirically-inductive sub-circle. Another 10 publications describe individual conceptions or approaches on how RPA and ML can be combined.

For more information concerning the research process, please see the available dataset comprising the detailed literature search settings including exclusion criteria, a list of the reviewed literature and our concept matrix [23].

## 4 A Nascent Taxonomy of ML-RPA Integration and Use

### 4.1 Overview

In total, we identified 8 dimensions with 24 characteristics. Table 1 presents the results in an aggregate form. The table is organized as follows: the first column contains our meta-characteristics, while the second column provides the associated dimensions. We documented the frequency of mentions of these dimensions in the literature in the third column. The fourth column contains our characteristics. The literature under consideration was initially categorized into the three types practical reports, conceptions, and frameworks. In the four columns on the right, we recorded the frequency of each characteristic in the respective publication type and as a total.

We will delve into our specific findings for each dimension in the subsequent sections, providing a detailed exploration of their individual characteristics and a comprehensive analysis of their occurrences. According to Nickerson et al. [29], a taxonomy with 8 dimensions is sufficiently differentiated yet concise. It thoroughly explains the differences in comparison to individual literature bases and can also be extended to include additional characteristics, such as other process environments.

| MC | Dimensions | M | Characteristics | # 45 | P 9 | C 10 | F 26 |
|---|---|---|---|---|---|---|---|
| RPA-ML integration | **Architecture and ecosystem** | 2 | External integration | 13 | 5 | 7 | 1 |
| | | | Integration platform | 11 | 2 | 2 | 7 |
| | | | Out-of-the-box (OOTB) | 3 | 1 | 0 | 2 |
| | **Capabilities of ML** | 4 | Computer vision | 21 | 6 | 1 | 14 |
| | | | Data analytics | 13 | 3 | 5 | 5 |
| | | | Natural language processing | 22 | 3 | 4 | 15 |
| | **Data basis for ML** | 1 | Structured | 2 | 1 | 0 | 1 |
| | | | Unstructured | 25 | 8 | 4 | 13 |
| | | | UI logs | 6 | 0 | 5 | 1 |
| | **Intelligence level** | 8 | Symbolic | 13 | 1 | 1 | 11 |
| | | | Intelligent | 33 | 8 | 3 | 22 |
| | | | Hyperautomation | 22 | 0 | 8 | 14 |
| | **Technical depth of integration** | 1 | High code | 11 | 5 | 5 | 1 |
| | | | Low code | 10 | 4 | 2 | 4 |
| RPA-ML interaction | **Deployment area** | 1 | Analytics | 7 | 1 | 2 | 4 |
| | | | Back office | 17 | 8 | 4 | 5 |
| | | | Front office | 7 | 0 | 3 | 4 |
| | **Lifecycle phase** | 1 | Process selection | 8 | 0 | 2 | 6 |
| | | | Robot development | 10 | 1 | 5 | 4 |
| | | | Robot execution | 18 | 8 | 2 | 8 |
| | | | Robot improvement | 5 | 1 | 1 | 3 |
| | **User-robot relation** | 5 | Attended | 12 | 5 | 4 | 3 |
| | | | Unattended | 10 | 4 | 2 | 4 |
| | | | Hybrid | 4 | 0 | 2 | 2 |

**Table 1.** Taxonomy of machine learning in intelligent robotic process automation. Legend: MC meta-characteristics, M mentions, # total, P practitioner reports, C conceptions, F frameworks

### 4.2 Architecture and ecosystem

First, we can make a distinction in how RPA software is architecturally and conceptually linked with ML capabilities. The question arises of whether ML components are distributed as separate parts of RPA software by the RPA provider, if RPA software opens up to become open an platform and ecosystem for third-party providers, or even adopts open-source models [10]. Alternatively, they can offer interfaces for individual external integration with ML. While frameworks particularly emphasize the platform approach ($n = 7$), practical reports still predominantly favor external linkage ($n = 5$).

**External integration.** Users can develop and train ML models themselves. Connecting to these models can be done through standardized APIs, but it requires programming effort and specialized knowledge for ML model development.

**Integration platform.** In this approach, RPA software transforms into an execution platform or engine [25]. This approach involves not only ML, but also additional modules for the direct integration of applications and services, eliminating the need for UI interaction or API-based programming [18]. The RPA software opens up to third-party specialized AI providers and open-source ML libraries, aligning with the rise of artificial-intelligence-as-a-service (AIaaS). AIaaS provides AI capabilities, including pre-trained models, through cloud-based training and configuration environments [10]. External ML capabilities can be integrated as offered modules into workflows. This approach allows for quick adoption of new ML trends and flexible module licensing.

**Out-of-the-box (OOTB).** This category encompasses integrations where ML capabilities are either built directly into the RPA software or can be added later via a robot store controlled by the software provider [33]. Crucially, the availability of ML capabilities depends entirely on the provider's offering.

### 4.3 Capabilities of ML

Four of the examined frameworks differentiate ML integration based on the specific ML capabilities utilized. However, most sources remain at a relatively high-level perspective without delving into technical details. According to Nickerson et al. [29], characteristics within a taxonomy dimension should ideally be mutually exclusive. Nevertheless, it has proven challenging to determine characteristics in this dimension in such a way that an approach can be unambiguously assigned to only one capability while not unbalancing our taxonomy by introducing excessively many characteristics. Given the context of *multi-skilling* [25], it is worth questioning whether intelligent robots might possess multiple ML capabilities and could thus be characterized by the combination of these capabilities. Further, some publications explicitly mention domains but do not specify the underlying algorithmic types. It is a dimension particularly affected by contemporary changes in AI capabilities due technological advancements.

**Computer vision (CV).** Here, we subsume all skills related to the capture and analysis of images in the broadest sense, such as optical character recognition (OCR) or intelligent character recognition (ICR), or generally, CV. In the case of OCR and ICR, the goal is to create machine-readable documents with a text layer from document images. While OCR is limited to printed documents, ICR can also recognize handwriting, for example. According to Kanakov and Prokhorov [20], approximately 90 % of all use cases of AI-augmented RPA use these techniques for document processing. In fact, 17 out of the 45 publications mention the possibility of using OCR. Among the 8 practical reports, OCR is used in 5. Outside the context of documents, ML models are used to recognize faces, classify images based on their content, pre-process text recognition, and also identify UI elements [34,20,41]. In the frameworks, CV, along with OCR, is the second most commonly mentioned capability.

**Data analytics.** In this case, ML does not serve to unlock data that would otherwise be inaccessible for automation, but rather facilitates the processing of data that is already usable. A common theme is the use of ML-based classification. This encompasses all abilities that classify data into various categories based on a previously trained ML model, serving, for example, in the preprocessing of emails or other documents such as splitting and prioritizing inquiries [25]. In this characteristic, we also capture models that, for example, aim to recognize and extract patterns from log files based on unsupervised learning.

**Natural language processing (NLP).** The use of ML has two perspectives in the domain of natural language. Traditionally, ML models in the sub-discipline of NLP are designed to process natural language texts and automatically extract their meaning [38]. NLP can be used for example not only in the execution phase to categorize user inquiries based on user sentiment, but it can be used also in other phases such as process selection for identifying automation candidates [26]. Other concepts involve configuring entire robots through natural language interaction [6]. In 22 out of 45 publications, the use of NLP in conjunction with RPA is considered, and in 3 out of 9 practical reports, NLP is included. Currently, due to the recent emergence of generative AI [11], the second perspective, namely natural language generation, is of significance. The integration of conversational agents in RPA primarily aims at enabling users, both end customers and employees, to interact with software robots in natural language, thereby making it as accessible as possible [33].

The use of natural language generation is closely intertwined with natural language understanding, where, for example, a three-part architecture initially uses NLP to determine the user's intent, then implements the user's request with RPA, and finally generates an appropriate response to the user, either rule-based or through a large language model (see *understand - act - respond*, [32]). Through the integration with RPA, a conversational agent can transform from a simple question-answer machine into a comprehensive assistant by triggering actions or transactions in existing application systems. By bridging the gap between the input and output of the conversational agent, the respective limitations are addressed as RPA can only process structured data into structured results, while conversational agents handle unstructured inputs leading to their natural language responses [8]. In our literature review, we found that connections to conversational agents are currently almost exclusively found in conceptual publications.

### 4.4 Data basis for ML

Since ML is strongly dependent on the underlying data, it is important to note which data is used in the context of RPA for both learning and applying a model.

**Structured.** RPA can retrieve data for the ML model, for instance, from legacy systems through the UI, making it usable for ML. The results of the ML model application can then be further processed based on rules. An example is insurance fraud detection, where RPA first extracts data from an Excel sheet, then the data records are classified by a pre-trained model, and depending on the results, various process variants are executed by RPA [30]. This way, ML can not just overcome the limitations of RPA, but RPA can also address the issue of data availability for ML.

**Unstructured.** Documents, video and audio files, as well as emails, need to be processed in 95% of all companies; 80% of the generated data is said to be semi- or unstructured [2]. However, RPA cannot process them in a rule-based manner without ML capabilities. ML makes the data usable for automation [41]. In the practical reports, the unlocking of unstructured data clearly dominates the other characteristics of this dimension.

**User interface logs.** To automate RPA itself, the system must learn autonomously from user interactions. In this case, ML models are not applied to external structured or unstructured data but rather to internal log files that would otherwise not be processed, or to representations of the robots themselves [33]. This characteristic is mainly observed in advanced conceptual approaches, while no practical report mentions learning from UI logs.

### 4.5  Intelligence level

The conceptual classification based on intelligence levels is the most common approach in the literature, appearing in various forms ($n = 8$). Other terms used for this classification are *automation types* or *stages* [7]. These classifications differentiate how intelligent the automation artifact is and how intelligent it behaves. These types are hierarchical, with higher stages expanding the capabilities of lower stages. Further proposed three-level classifications, such as the automation waves according to PricewaterhouseCoopers International, can be recognized as strongly related to this dimension [39]. Table 1 indicates that practical reports are almost exclusively associated with the stage *intelligent*, while frameworks and conceptions also focus on *hyperautomation*.

**Symbolic.** This stage includes traditional rule-based RPA, as described in Section 2. Systems in this category rely on structured data and can only perform tasks for which they were deterministically configured [7]. The corresponding automation wave is the *algorithm wave*, where simple arithmetic and data analysis tasks are automated [39]. Basic Automation operates relatively mechanically, replacing the "arms and legs" of human workers [41]. Robots in this category are referred to as *doing bots*, indicating that they perform routine tasks and are otherwise dependent on humans.

In analogy to symbolic AI, Herm et al. [17] refer to this stage as *symbolic* RPA because it faces similar constraints as knowledge-based and expert systems in the AI domain. Symbolic RPA robots can only handle tasks for which they have been explicitly programmed or configured, but complex and cognitive tasks are challenging to formalize in this way.

Nevertheless, even in symbolic stage tools without externally noticeable intelligent features, ML techniques are being used to adapt, for example, to changing UI surfaces [41]. To reflect this internal usage in our taxonomy, we have decided to include this characteristic.

**Intelligent.** In this second stage, robots can perform more complex processes by incorporating specific cognitive abilities. This includes processing unstructured or semi-structured data. Robots in this stage are referred to as *thinking bots*. They can, for example, process natural language because they have been trained for this specific task [7]. These robots are specialists in their respective domains and problems and are limited to those areas. They cannot autonomously transfer their skills to other tasks, learn new things, or self-improve. This corresponds to the concept of narrow AI [10]. Nevertheless, CA and intelligent automation (IA) no longer operate strictly deterministically but with well-defined probabilistic components. The execution logic remains rule-based to remain predictable and meet manually specified requirements [41]. The corresponding automation wave is the *augmentation wave*, predicted by PricewaterhouseCoopers International for the late 2020s [39]. The term *intelligent robotics* also falls into this category [36].

**Hyperautomation.** In advanced automation, robots learn adaptively from data and experience, and they manage and improve autonomously. These robots are referred to as *learning bots* [7] or autonomous agents [28]. *Hyperautomation* is also part of this type. The Greek prefix "hyper" implies that automation occurs at a higher level. It is about automating the process of automating individual processes through a higher-level perspective, thus improving it [17]. Hyperautomation aims to automatically generate suitable automation artifacts such as scripts, robots, or workflows [35]. The corresponding automation wave here is the *autonomy wave* predicted for the mid-2030s, where problem-solving is automated [39]. Currently, there is no agreed upon definition and scope of hyperautomation as it originated from recent business practice. Its detailed specification and appearance will emerge in the coming years.

### 4.6   Technical depth of integration

Closely tied to the topic of architecture is the question of the technical depth into which an RPA developer must delve to integrate ML in RPA software. Consequently, this dimension depends to some respect on the architecture on the provider side, but takes the complementary perspective of the users. Agostinelli et al. [1] distinguish between "strong coding", graphical user interface (GUI), and low-code tools. Due to the minimal difference between GUI (viz. no code) and low code, we only differentiate between high code (viz. classical formal language programming) and low code. While low-code paradigms clearly dominate in frameworks ($n = 4$ vs. $n = 1$), in the examined practical reports, no paradigm has clearly prevailed ($n = 4$ vs. $n = 5$).

**High code.** Integrating ML capabilities to RPA robots requires programming in high-level, domain-specific programming or scripting languages. The skills of the integration experts would need to be those of a programmer rather than of a user.

**Low code.** No-code or low-code environments do not require knowledge of programming languages as robots are generated through UI-based modeling. The skills of the person responsible for the integration is that of a power user and may be available in business departments rather than the IT department.

### 4.7 Deployment area

Integrated RPA and ML solutions are opening up to process tasks from the back office up to the front office as technology advances [8]. Rather than automating transactional tasks intelligently, a strong focus on intelligent automation of decisions can be observed. With these three characteristics, we try to subsume these developments.

**Analytics.** AI-enhanced RPA robots can make decisions more flexibly than rule-based robots and are faster and more consistent than humans. The decision-making is informed based on a data foundation, but it is not without its challenges. These processes are no longer repetitive and straightforward tasks but have become so complex that they would typically be handled by experts [22]. Consequently, rules-based approaches do not suffice.

**Back office.** Even though structured data is no longer mandatory, simple, structured, and matured processes in back-office contexts remain highly suitable for RPA [41]. In fact, [20] state that 90% of intelligent RPA applications would still perform back-office tasks, which can now also be based on unstructured documents through OCR and intelligent document processing as well. This is supported by 8 out of 9 practical reports that can be attributed to this characteristic.

**Front office.** For example in combination with conversational agent technology, RPA can be used not only in the back office but also to interact directly with customers. Automation enables faster customer service and more availability [8].

### 4.8 Lifecycle phases

ML techniques are employed at various stages when creating, using, and improving RPA robots. These phases are distinct from the project management phases of an RPA project, as outlined by [16]. The explicit distinction among these phases is evident in the framework proposed by Agostinelli et al. [1].

**Process selection.** Before creating an RPA robot, it is crucial to identify which routines to automate. ML techniques can automatically detect these tasks. Typically, process selection relies on manual methods like interviews, observations, or document analysis, which can overlook unconscious routines and involve significant effort [9]. Alternatively, process mining or task mining can be used, involving the use of UI logs to automate process discovery and data-driven analysis [1]. This is referred to as *inter-routine self-learning*, where various routines with automation potential are autonomously identified. However, the subsequent handling of these routines traditionally requires implementation by human experts.

**Robot construction.** In standard RPA software, robots are created in low-code environments by developers who build models based on recorded UI interactions [33]. Automating this process would require RPA software to autonomously derive rules from user observations to instantiate a robot [12]. The challenge lies in operational user interaction data, which may contain non-linear, interchangeable routines and disruptive noise. Automatic segmentation is difficult, especially as UI logs typically lack a label (or case ID) for the underlying business case. This type of learning is termed *intra-routine self-learning* [1]. For instance, [37] suggest using shallow ML techniques to segment similar operation sequences into potential routines. To make a robot ready for operation, the actual model (i.e., flowchart) must be constructed from analyzed user interaction data. User-driven flowchart construction poses risks, particularly in complex processes, as it may lead to faulty robots interacting with production systems and causing harm due to human errors or the neglect of exceptional cases, mainly due to a lack of testing environments [1].

**Robot execution.** In the examined publications describing real-world applications, ML is used in 8 out of 9 cases during the execution phase of the robots. This application does not improve the automation process itself but enhances the robot's task execution. Pre-trained models are employed to process unstructured data such as documents, or conversational agents are used to interact with customers using natural language. Moreover, in extensive RPA application scenarios, complex sequences of interdependent routines are automated. Concepts of automatic planning can be used to automatically determine intelligent execution strategies [1,41].

**Robot improvement.** In RPA software, even without apparent intelligent features, CV techniques are often integrated to tackle the challenge of evolving UI interfaces. They serve to ensure the continued identification of UI elements, particularly through virtual interfaces like Citrix [38,41]. More advanced concepts propose the use of experiential learning from mistakes during this phase, aiming to generalize individually created but similar robots. This addresses the lightweight nature of RPA, which tends to lead to localized and bespoke solutions. To reintroduce standardization across organizational processes, robots with similar tasks should be merged and harmonized [33].

### 4.9  User-robot relation

This dimension focuses on the role of humans during the robot's process execution. It distinguishes how humans interact with the robot and in what environment the robot operates, whether in desktop environments of workstations or on central servers. This distinction is already made for classical RPA [7] and does not directly characterize the connection with ML. However, ML influences which interactions with humans are still necessary, leading to shifts in this dimension. That is why we included it in our taxonomy. In the literature, no interaction form clearly dominates across all types of publications. Hybrid approaches are only mentioned in frameworks and concepts. Therefore, we presume that ML components enable various forms of interaction and overcome previously existing limitations.

**Attended.** This characteristic centers around the terms *desktop-level* automation and *human-in-the-loop*. RPA is triggered by humans for specific tasks and constantly interacts with the user on their individual desktop [7]. RPA is considered a tool for the individual user [39].

**Unattended.** RPA is executed on central servers and runs continuously in the background without human intervention. This is why it is also referred to as *enterprise-level* automation [7]. Humans are only required for handling exceptional cases [39]. In recent years, RPA has been shifting away from its original focus of desktop automation towards this level [21].

**Hybrid.** Due to the different forms of intelligence possessed by humans and machines, the connection towards a collaborative partnership between robots and users is seen as an enrichment. While humans can use social and experiential knowledge and possess creativity, intuition, and empathy, the "intelligent" robot can quickly, efficiently, and consistently process available data analytically based on underlying probability distributions [10,39].

## 5  Application, Limitations, and Summary

In our work, we explored the possibilities of integrating the two concepts of RPA and ML to mitigate the limitations inherent to symbolic RPA. The goal was to systematize the theoretical research landscape strewn across both communities which resulted in a nascent taxonomy comprising 2 meta-characteristics with 8 different dimensions that describe the connection of RPA and ML. Despite the high relevance of our literature corpus, we must take into account that the market develops rapidly, with recent annual growth rates ranging from 17.5% to 30.9% in the years 2021 to 2023 [13]. This leads to very dynamic developments in the RPA software products available on the market, which are increasingly equipped with more intelligent features – especially in their advertising.

**Application and evaluation.** Hence, in a bid also to evaluate the usefulness and applicability of our literature-driven taxonomy, we have reviewed the current products of *UiPath*, *Automation Anywhere*, and *Microsoft Power Automate* that are leading Gartner's magic quadrant for RPA [14]. The evaluation took place from April to June 2023. We used the community editions where available and employed the taxonomy as an inspection catalog. We did not encounter any substantial concepts (viz. software functionality) that we could not classify with our taxonomy.

Our comparison of the RPA products showed that there are major similarities between all three RPA products in terms of similarities in the range of functions, the structure in similar modules, and the focus on low code. For example, all software vendors rely on a platform approach for not only ML components, but also for the integration of other APIs in marketplaces managed by them. At the same time, all RPA products offer their own ML components, especially for document processing. On all platforms, not only desktop automation, but also cloud automation via APIs is possible. The use of all capabilities of ML as mentioned in the taxonomy is either available OOTB or by third-party providers. All RPA products focus their ML components primarily on the phases of process selection (through task mining) and process execution. For ML-supported design, Automation Anywhere and Microsoft Power Automate have each recently started to offer copilot functions based on conversational agents (i.e. chatbots). However, we could not find any ML-based automatic improvement functions. Thus, we classify the RPA products as IA, but they obviously market as and aim for hyperautomation functionality. Yet, autonomous learning and autonomous adaptation are nowhere to be found as of today.

**Limitations.** As with all research, our taxonomy has some limitations. First, we did not perform an exhaustive structured literature review, but we opted for a hermeneutic approach to better understand the intricacies of evolving topic of RPA-ML integration. Thus, we stopped at a point of saturation to offer a nascent taxonomy for discussion rather exhaustive taxonomy that cannot yet be complete as ML continues to advances while software engineers are busy implement yesterday's advances into their products. Also, we must recognize that there are interdependencies between some of our dimensions that cannot be resolved easily: for example, the architecture significantly influences the technical depth of integration. Different levels of intelligence require varying capabilities of ML. Depending on the process context and the data being processed, various capabilities of ML from different domains are required. That is, dimensions are closely interconnected and should not be considered in isolation. Further, our capabilities dimension resides at a high level and could be detailed into concrete CV, NLP, and data analytics techniques. For the moment, we have consciously decided against detailing this dimension as it would significantly unbalance the taxonomy with respect to the other dimensions' level of detail.

**Discussion and summary.** Summarizing, we found that practical applications of RPA-ML integration are primarily attributed to the intelligence level of *intelligent* RPA, which means that they work with specialized cognitive capabilities, especially in the execution phase. However, there is an effort to focus on the entire automation process and to make automation itself smarter through the application of ML, alluding to the still underspecified concept hyperautomation promising self-learning and generative robots that scale as the tasks require.

In addition, we observed that in practical reports, ML is primarily used in the execution phase with a wide range of ML capabilities. This also corresponds to the level of intelligence of *intelligent*. As narrow but superintelligent *thinking bots*, RPA robots can perform specialized tasks, such as processing unstructured data, on which they have been explicitly trained. Academic concepts predominantly focus on *hyperautomation* and self-learning robots.

Further, with the advent of generative AI we see a shift towards the process selection phase in particular. Driven by recent successes, software vendors are beginning to incorporate co-pilot functionality for the specification or automated construction of RPA robots in their products. However, contrary to some concepts from literature, human interaction is still very much necessary for this endeavor, as RPA software does not (yet) independently derive rules or analytical models from pure user observation.

The landscape of intelligent RPA is bound to evolve further especially with the recent advent of generative AI. Hence, our nascent taxonomy will require constant updates as technology advances. Nevertheless, our taxonomy can already be used today to generate archetypes of intelligent RPA robots and assist organizations in comparing the promises of next-generation RPA products in the market to assist the design of complex intelligent RPA building blocks and patterns such as those aforementioned neutral data trust models that go beyond the capabilities of symbolic RPA.

## Acknowledgements

This research and development project is funded by the German Federal Ministry of Education and Research (BMBF) within the "Richtlinie zur Förderung von Projekten zur Erforschung oder Entwicklung praxisrelevanter Lösungsaspekte ("Bausteine") für Datentreuhandmodelle" (Funding No. 16DTM201B) and financed by the European Union - NextGenerationEU. The authors are responsible for the contents of this publication.

Taxonomy of Intelligent Robotic Process Automation    17